\documentclass[preprint,12pt]{elsarticle}

\usepackage{amssymb}
\usepackage{siunitx}
\usepackage{amsmath}
\usepackage{hyperref}
\usepackage{fancyhdr}
\usepackage{multirow}
\usepackage{makecell}
\usepackage{rotating}
\usepackage{lineno}
\usepackage{xcolor}  % for color

\usepackage{pdflscape}

\journal{Journal of Manufacturing Processes}

\begin{document}

\begin{frontmatter}


\title{QA-VLM: Providing human-interpretable quality assessment for wire-feed laser additive manufacturing parts with Vision Language Models}

\author[inst1]{Qiaojie Zheng}
\author[inst2]{Jiucai Zhang}
\author[inst1]{Joy Gockel}
\author[inst3]{Michael B. Wakin}
\author[inst1]{Craig Brice}
\author[inst1]{Xiaoli Zhang\corref{cor1}}

\affiliation[inst1]{organization={Mechanical Engineering, Colorado School of Mines},%Department and Organization
            addressline={1500 Illinois St.}, 
            city={Golden},
            postcode={80401}, 
            state={CO},
            country={USA}}
\affiliation[inst2]{organization={GAC R\&D Center Silicon Valley},%Department and Organization
            city={Sunnyvale},
            postcode={94085}, 
            state={CA},
            country={USA}}
            
\affiliation[inst3]{organization={Electrical Engineering, Colorado School of Mines},%Department and Organization
            addressline={1500 Illinois St.}, 
            city={Golden},
            postcode={80401}, 
            state={CO},
            country={USA}}

\ead{xlzhang@mines.edu}
\cortext[cor1]{Corresponding author.}

\begin{abstract}
Image-based quality assessment (QA) in additive manufacturing (AM) often relies heavily on the expertise and constant attention of skilled human operators. While machine learning and deep learning methods have been introduced to assist in this task, they typically provide black-box outputs without interpretable justifications, limiting their trust and adoption in real-world settings. In this work, we introduce a novel QA-VLM framework that leverages the attention mechanisms and reasoning capabilities of vision-language models (VLMs), enriched with application-specific knowledge distilled from peer-reviewed journal articles, to generate human-interpretable quality assessments. Evaluated on 24 single-bead samples produced by laser wire direct energy deposition (DED-LW), our framework demonstrates higher validity and consistency in explanation quality than off-the-shelf VLMs. These results highlight the potential of our approach to enable trustworthy, interpretable quality assessment in AM applications. 
\end{abstract}

\begin{keyword}
Intelligent Manufacturing \sep Laser-Wire Direct Energy Deposition \sep Vision Language Models \sep Quality Assessment

\end{keyword}

% \begin{keyword}
% %% keywords here, in the form: keyword \sep keyword
% Laser Powder Bed Fusion \sep Transfer Learning \sep Process-Property Mode \sep Active Learning
% %% PACS codes here, in the form: \PACS code \sep code
% % \PACS 0000 \sep 1111
% %% MSC codes here, in the form: \MSC code \sep code
% %% or \MSC[2008] code \sep code (2000 is the default)
% % \MSC 0000 \sep 1111
% \end{keyword}

\end{frontmatter}


% \begin{linenumbers}
\section{Introduction}\label{Intro}

Laser wire direct energy deposition (DED-LW) is a near-net-shape metal additive manufacturing process suitable for producing a wide range of materials\cite{Marini2017, Kumar2020}. It has shown significant promise in the fabrication of large-scale components with high deposition rates and reduced material waste~\cite{Feier2022, Cunningham2017}. These advantages have made DED-LW especially attractive for industries that require a high deposition rate to make large parts~\cite{Wu2018, Li2022}. DED-LW has been successfully employed in producing aerospace components with complex geometries and tailored mechanical properties~\cite{Omiyale2022} and in producing marine structures where corrosion resistance and scalability are essential~\cite{Choi2024, Vishnukumar2021}. Despite its great potential, the complex production process often makes it difficult to tune process parameters to produce parts with predictable mechanical properties~\cite{Vilaro2012, Wang2022}. As a result, the DED-LW parts often need to be examined to verify the properties. Visual examinations, including assessments of surface roughness, contour smoothness, and cross-sectional geometries, are required to provide valuable insights into the part’s quality.

\begin{figure}
    \centering
    \includegraphics[width=0.75\linewidth]{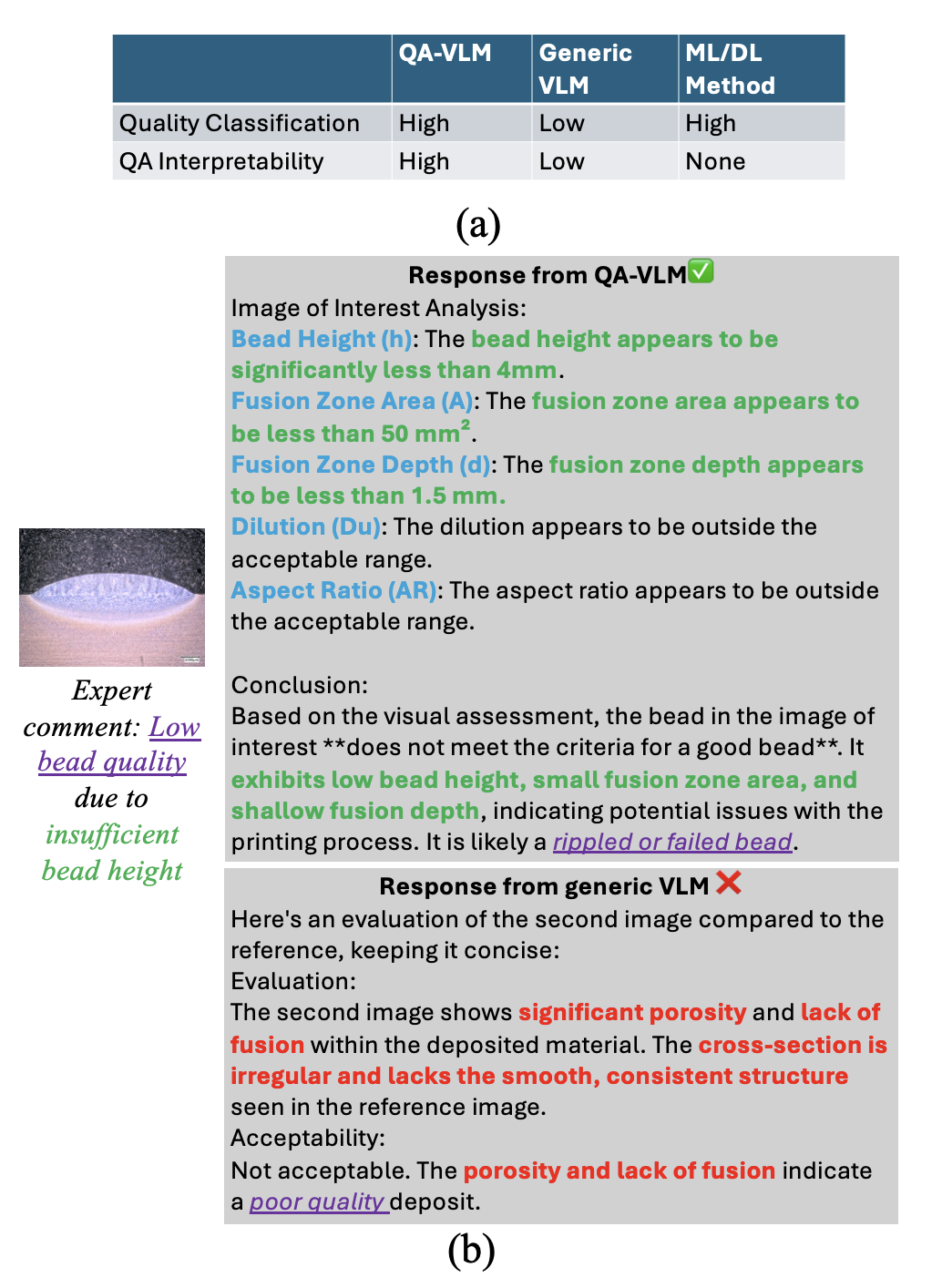}
    \caption{(a) Model performance and interpretability comparison among the introduced QA-VLM, Generic VLM, and existing ML and DL approaches. (b) Illustration of the correct assessment from QA-VLM and the incorrect assessment made by the generic VLM. Gemma-3 was used in both cases. Although both models identified the bead as a portion of a poor-quality print (marked by italic purple font), responses from the generic VLM provided incorrect reasoning (marked by bold red font) as lack of fusion and porosities. On the contrary, the assessment response made by QA-VLM, powered by the same Gemma-3 model, contains correct reasoning (marked in bold green font) as undesirable bead geometries}
    \label{fig:Figure_1}
\end{figure}

Progress in data-driven modeling techniques, such as machine learning and deep learning (DL), over the last few decades has demonstrated the potential for automating quality assessment (QA) in visual examination for additive manufacturing (AM) applications~\cite{Zhou2021, Gobert2018, Garg2015, Meng2020}. These methods can directly extract and correlate features from sensing data, such as RGB imagery, thermal imagery, and photodiode signals, with the quality of printed parts. In applications to laser powder bed fusion, Scime et al.~\cite{Scime2020} developed an anomaly classification algorithm using k-means clustering. Layer-wise image data were processed through a series of computer vision filters to extract features, which were subsequently grouped to detect anomalies. This approach achieved a true positive rate of 89\% for anomaly detection, although classification accuracy varies depending on the anomaly type. In the domain of DED-WL, Liu et al.~\cite{Liu2022} employed a naïve Bayes classifier to predict overall bead quality based on process parameters. Their model categorized prints into three classes—smooth, rippled, and failed—achieving an accuracy of 73.74\%. Despite the great autonomous achievement brought by these methods, the greatest disadvantage of these ML and DL approaches is the lack of interpretability in their outputs~\cite{Mattera2024, Vozza2024, Ukwaththa2024}. These models typically operate as black boxes, constructing end-to-end mapping rules solely based on statistical patterns in the training data~\cite{Qamar2023, Hassija2023}. While their assessment conclusion may be accurate within the scope of their training, they often fail to explain the rationale behind their decision, as summarized in Figure~\ref{fig:Figure_1}(a). This lack of interpretability significantly undermines user trust, especially in scenarios involving domain shifts, such as changes in machines or feedstock suppliers, where the system behavior may deviate from the training distribution. Consequently, existing ML- and DL-based quality assessment models are not widely adopted in real production settings. In most cases, quality assessments still rely heavily on experienced human experts who can provide justifiable and interpretable evaluations. 

Recent advancements in large language models (LLMs) and vision-language models (VLMs), such as OpenAI’s GPT and Google’s Gemini, have opened new opportunities for making image-based quality assessment more interpretable and justifiable. By leveraging powerful attention mechanisms, extensive pre-trained knowledge, and strong image reasoning capabilities, VLMs have the potential to perform visual assessments in a human-like manner that provide not only the classification results but also the reasons to justify them. Despite the strong attention and reasoning capabilities, existing VLMs are not yet ready to be deployed for quality assessment tasks for additive manufacturing. Being general-purpose VLMs, these models are only trained on broad, publicly available datasets where detailed information about AM is scarce. Consequently, these generic models often exhibit inadequate understanding and reasoning capabilities when applied to metal AM tasks, leading to incorrect assessment conclusions and supporting reasons. For instance, in the bottom half in Figure~\ref{fig:Figure_1}(b) the model erroneously attributes white speckles in the black supporting material to lack of fusion, while the true attention and analysis should be placed on the bead cross section. Such misplaced attention and incorrect conclusion demonstrate the insufficient application-specific knowledge from the generic VLMs.

Due to this knowledge gap, current applications of LLMs and VLMs in AM are primarily confined to tasks like text-based information retrieval and summarization, such as in AMGPT~\cite{Chandrasekhar2024}, or diagnostics in plastic-based processes like Fused Deposition Modeling (FDM)~\cite{Jadhav2024}, where training data is more readily available. One promising path toward expanding the utility of these models in metal AM is domain-specific fine-tuning, which has proven effective in other technical fields~\cite{Lu2025}. However, fine-tuning requires significant investment in both data collection and computational resources, making it an impractical solution in many cases.

In this work, we present a novel QA-VLM framework aimed at addressing the knowledge limitations of VLMs in specialized domains. The framework pursues two key objectives: 1) equipping generic VLMs with application-specific knowledge by extracting textual descriptions from peer-reviewed journal articles, and 2) enabling these knowledge-enhanced VLMs to correctly apply the acquired knowledge to assess print quality from images with reference-guided knowledge grounding. Together, these two objectives enable the QA-VLM to produce a valid print quality assessment based on the cross-section image, as demonstrated in the top half in Figure~\ref{fig:Figure_1}(b).

To demonstrate the effectiveness of the framework, we applied the QA-VLM to a print quality assessment task involving 24 single-bead prints produced by the DED-LW process. To evaluate the framework’s performance across different VLM architectures, we included two backbone models, Gemini 2.0 Flash (Gemini) and Gemma-3 (Gemma), in this study. Gemini is a larger model with around 40 billion parameters to offer broader knowledge and better reasoning capability, while Gemma is a smaller model with around 27 billion parameters with less knowledge and reasoning capability. The domain-specific knowledge for quality assessment is sourced from one peer-reviewed journal article focused on the same topic. While the framework is capable of incorporating multiple studies to enrich the knowledge base, we limit it to a single source in this work to ensure processing efficiency, reduce variability in downstream comparisons, and clearly demonstrate core functionality. 

To further evaluate the contributions of each objective, we performed ablation studies by systematically removing components corresponding to each objective and assessing their impact on performance. The results from ablation studies demonstrate that the full two-objective framework yields the greatest performance improvement, though partial gains were also observed when each objective was implemented independently compared to the baseline off-the-shelf VLMs. 

In summary, our contributions are:
\begin{itemize}
    \item We present a two-objective QA-VLM framework that enables generic VLMs to apply domain-specific knowledge that is extracted from literature to provide human-interpretable print quality assessments based on images of single-bead DED-LW.
    \item We demonstrate the capabilities of the QA-VLM framework in identifying key characteristics, such as shallow fusion zone and low bead height, from sub-bar prints. 
    \item We conduct ablation studies by intentionally leaving out QA-VLM components to evaluate the individual and combined impact of each objective, highlighting the importance of their integration for optimal performance. 
\end{itemize}

\section{Methodology}
\begin{figure}
    \centering
    \includegraphics[width=1.0\linewidth]{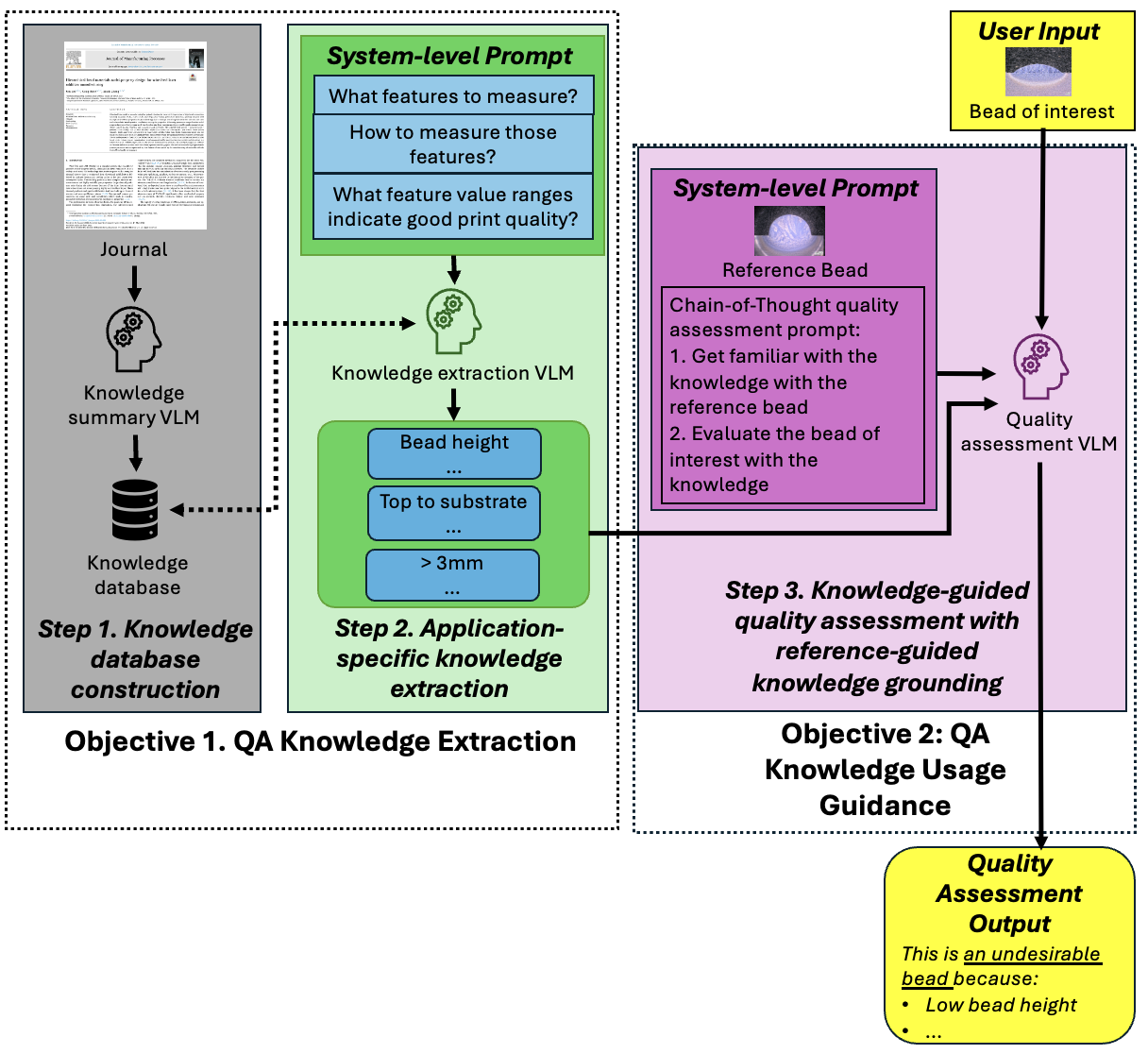}
    \caption{The two-objective QA-VLM framework lets the VLM make quality assessment (QA) based on application-specific knowledge. The first objective is to extract the text description of the knowledge specific to the QA task; the second objective is to let VLMs learn how to apply it. The first objective consists of two steps: constructing a knowledge database from journal articles and extracting knowledge related only to QA. The second objective involves a reference-guided knowledge grounding strategy that enables VLMs to learn from high-quality reference images. Together, these two objectives enable VLMs to output valid and well-justified outputs, rather than erroneous ones from only generic VLMs.}
    \label{fig:Figure_2}
\end{figure}

As outlined in Section~\ref{Intro}, the core concept of the QA-VLM framework is to fulfill two primary objectives: 1) bridging the knowledge gap between generic vision-language models (VLMs) and domain-specific tasks by extracting relevant information from peer-reviewed articles, and 2) enabling the appropriate application of this knowledge in the print quality assessment process. The first objective comprises two steps: constructing a knowledge database and retrieving application-specific knowledge. The second objective involves guiding the VLM to interpret the retrieved knowledge and apply it correctly during assessment. The overall workflow is illustrated in Figure~\ref{fig:Figure_2}, with detailed explanations provided in the following subsections. 

\subsection{Knowledge extraction}
The purpose of the two-step knowledge extraction is 1) to make an easy-to-search knowledge database by converting the knowledge in the journal articles from lengthy and unorganized descriptions into succinct and organized ones, and 2) to retrieve and distill application-specific knowledge for downstream applications and discard the irrelevant ones. 

\subsubsection{Knowledge database construction}
Directly retrieving knowledge from journal articles is challenging due to their length and complexity. To address this, each article is segmented into coherent sections, and each section is converted into a concise summary to enable faster and more efficient retrieval. These summaries are then vectorized and stored in a knowledge database, following the principles of the Retrieval-Augmented Generation (RAG) framework~\cite{lewis2020retrieval}, for downstream processes.

The summarization is performed by a summarization VLM, and the detailed steps are illustrated in Figure 3. A long article is first broken down into small semantically coherent chunks using the Unstructured library, and then each chunk is summarized based on the context of the article. The summarization process mimics human skimming behavior when reading journal articles. Rather than reading the entire text word-for-word, we typically first acquire a high-level understanding of the articles by reading the title, abstract, and conclusion, then selectively dive deep into sections of interest. Similarly, the knowledge summaries are acquired in a context-aware manner, where the summarization VLM is provided with context information acquired from the title, abstract, and conclusions. To provide appropriate summaries for non-text contents, such as tables and figures, the summarization VLM is also provided with their associated captions, titles, and nearby descriptive paragraphs. It should be noted that a human expert is required to select relevant journal articles during the initial setup of the knowledge database. However, their involvement is limited to this one-time framework configuration and is not needed for the ongoing assessment of individual print parts.

\begin{figure}
    \centering
    \includegraphics[width=1.0\linewidth]{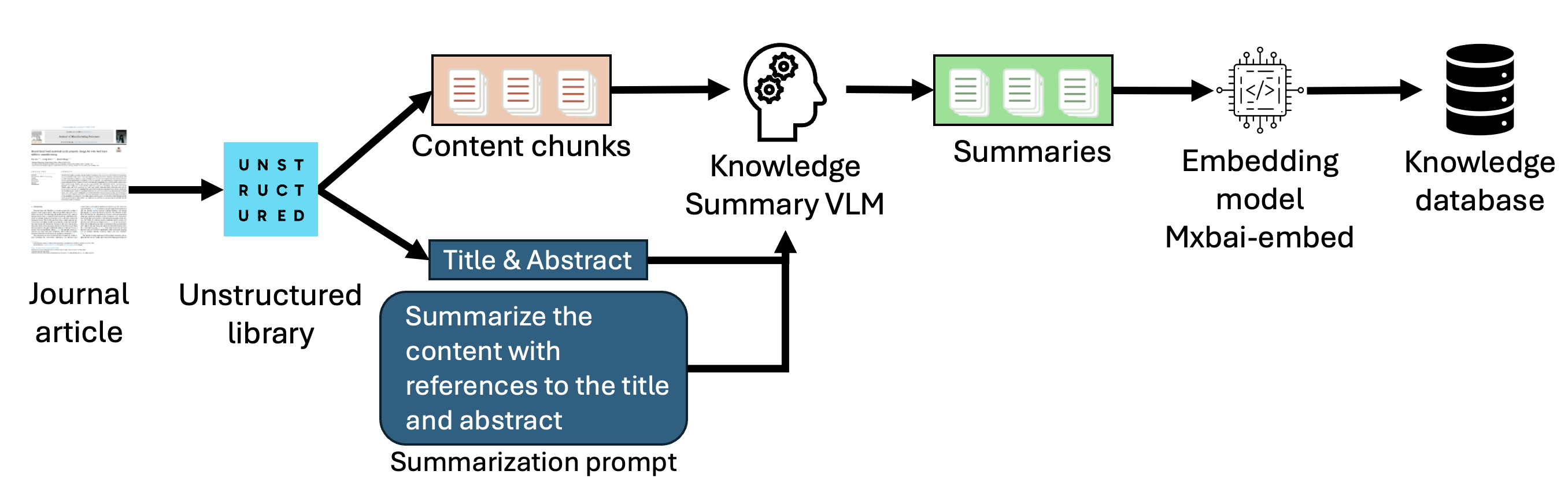}
    \caption{Process of knowledge database construction}
    \label{fig:Figure_3}
\end{figure}

The summarized knowledge descriptions should be succinct and informative enough to allow efficient retrieval. For building the database from multiple articles, the above steps can be repeated for each journal independently, and then the knowledge summaries can be pooled together. Since each summary is generated under a unique context, the summarized knowledge will not be confused with that of other contexts for downstream retrieval. Following the knowledge database construction step in RAG, these descriptions are ready to be vectorized. To achieve this conversion, the Mxbai-embed model~\cite{mxbai2024embed} is used to convert short descriptions of knowledge into a 1024-dimensional vector, which is ready for efficient retrieval. Because this knowledge database construction does not rely on domain-specific prompts, this process may also be generalized to other areas that involve quality assessment.

\subsubsection{Application-specific knowledge extraction}\label{application specific knowledge extraction}
Efficient knowledge retrieval becomes feasible once the database has been constructed. Although the database constructed from the journal article is highly related to the quality assessment task, not all summaries in the knowledge database are useful. The database could still contain much irrelevant information that could cause confusion or inaccuracies in the quality assessment step. Therefore, the goal of this step is to extract or distill the knowledge relevant only to the quality assessment application and discard the rest. We break down the quality assessment into a three-step reasoning process, for which the corresponding needs to be extracted from the database. The key pieces of knowledge involved in these steps are: 1) What features correlated to quality need to be extracted? 2) How to extract those features from images? and 3) What are the value ranges for the features to be considered good?

\begin{figure}
    \centering
    \includegraphics[width=0.6\linewidth]{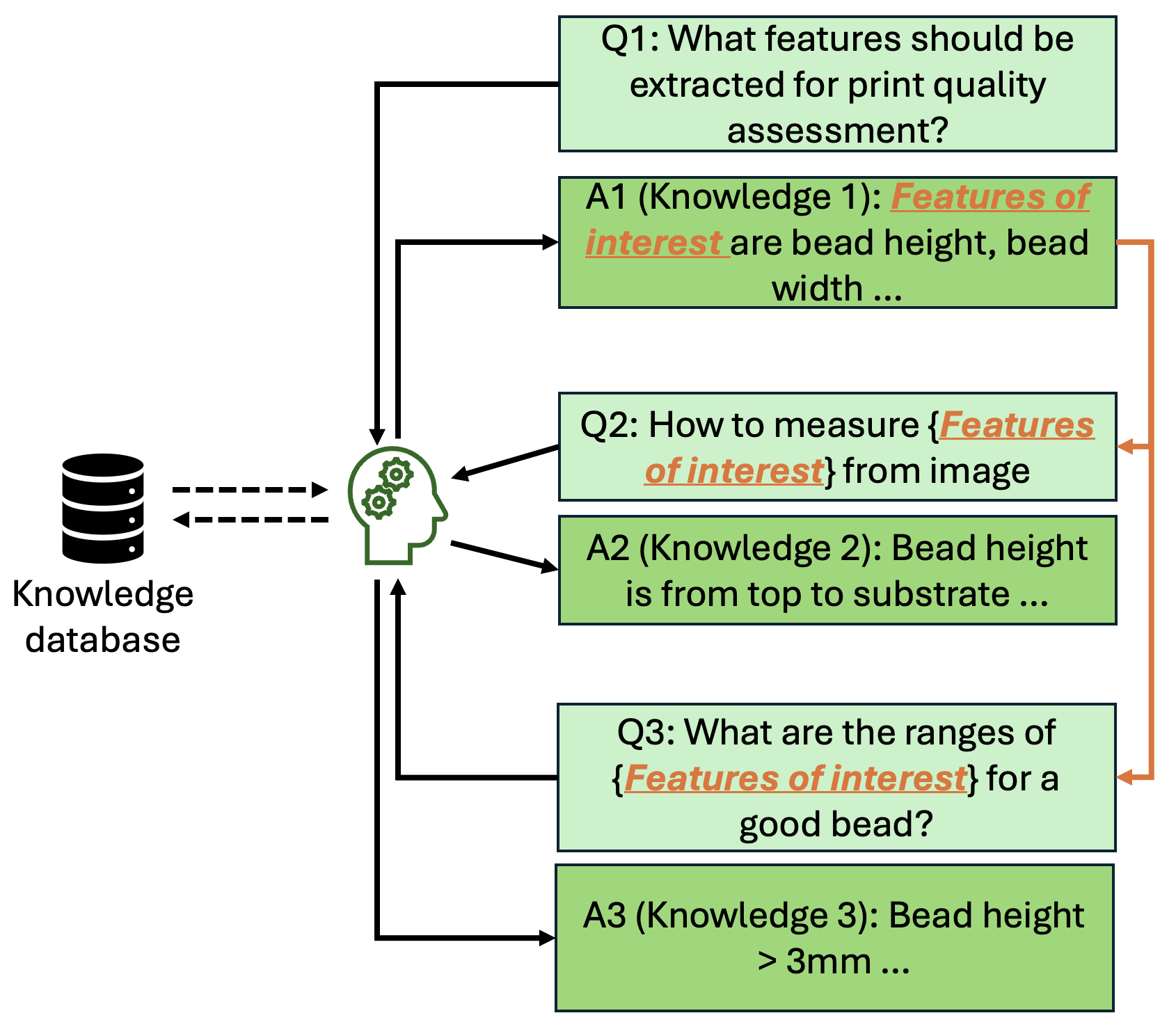}
    \caption{Knowledge extraction process to acquire important features to measure, how to measure them, and the range for good quality}
    \label{fig:Figure_4}
\end{figure}

Following this reasoning process, knowledge extraction is formulated as a sequence of interdependent queries to a knowledge extraction VLM to seek the most relevant knowledge descriptions in the database. The first query identifies the features of interest, which then inform subsequent queries. For example, an initial question could be: “What features are commonly used to distinguish good and bad prints in DED-LW manufacturing?” Once the features of interest are identified, the subsequent queries about feature measurement and good-value regions can be carried out. Since these two queries do not have interdependency, they can be carried out in parallel. Suppose that the query about the feature of interest returned the answer of “bead width”, and an example for the remaining two questions could be “how to measure bead width for DED-LW” and “what range of bead widths are considered good for DED-LW”. The overall process is illustrated in Figure~\ref{fig:Figure_4}. Although DED-LW is used here as an illustrative context, this framework is adaptive and can be applied to other manufacturing processes by adjusting the domain-specific context of queries. Human experts may be required in this knowledge extraction step to set up efficient queries, but once again, human effort is only required during initial configuration and is not required in the assessment for individual parts. 

The knowledge extraction procedure will first search for the top few summaries that are most relevant, i.e., most semantically similar, to the input questions. For example, when asked with the first query about what feature to use, this process should return summaries that talk about the feature types to be included, rather than detailed ranges of features. To achieve this goal, we convert the query into number embeddings with the same embedding model and look for the summaries whose embeddings in the knowledge database are the closest to the query input. Formally, for a knowledge database $D$, an embedding model $\phi$, and the query question $x$, the top $N$ relevant knowledge $K(x)$ summaries are obtained with Equation~\ref{eqn:eqn1} by looking for those knowledge embeddings that have the highest cosine similarity with the query embedding. These retrieved knowledge summaries are then passed into a VLM for summarization into concise responses for downstream use. 

\begin{equation}\label{eqn:eqn1}
K(x)=\left\{k_1, k_2, \ldots, k_N\right\} \text { where } k_i \in \arg \max \cos (\phi(x), \phi(k))
\end{equation}

\subsection{Reference-guided knowledge grounding} \label{knowledge grounding}
With the key knowledge in properties, procedures, and value ranges identified in Section~\ref{application specific knowledge extraction}, we are ready to equip generic VLMs with a textual description of quality assessment knowledge. However, these knowledge-equipped VLMs often fail to produce reliable answers because they have not grounded their understanding of the textual descriptions in the visual characteristics of real images from the DED-LW process. In other words, these models lack the ability to appropriately apply the supplied domain knowledge to actual visual data during assessment. For instance, a VLM might incorrectly measure bead height as the distance from the top of the bead to the bottom of the heat-affected zone, rather than the correct definition, which is from the top of the bead to the surface of the substrate.

\begin{figure}
    \centering
    \includegraphics[width=1.0\linewidth]{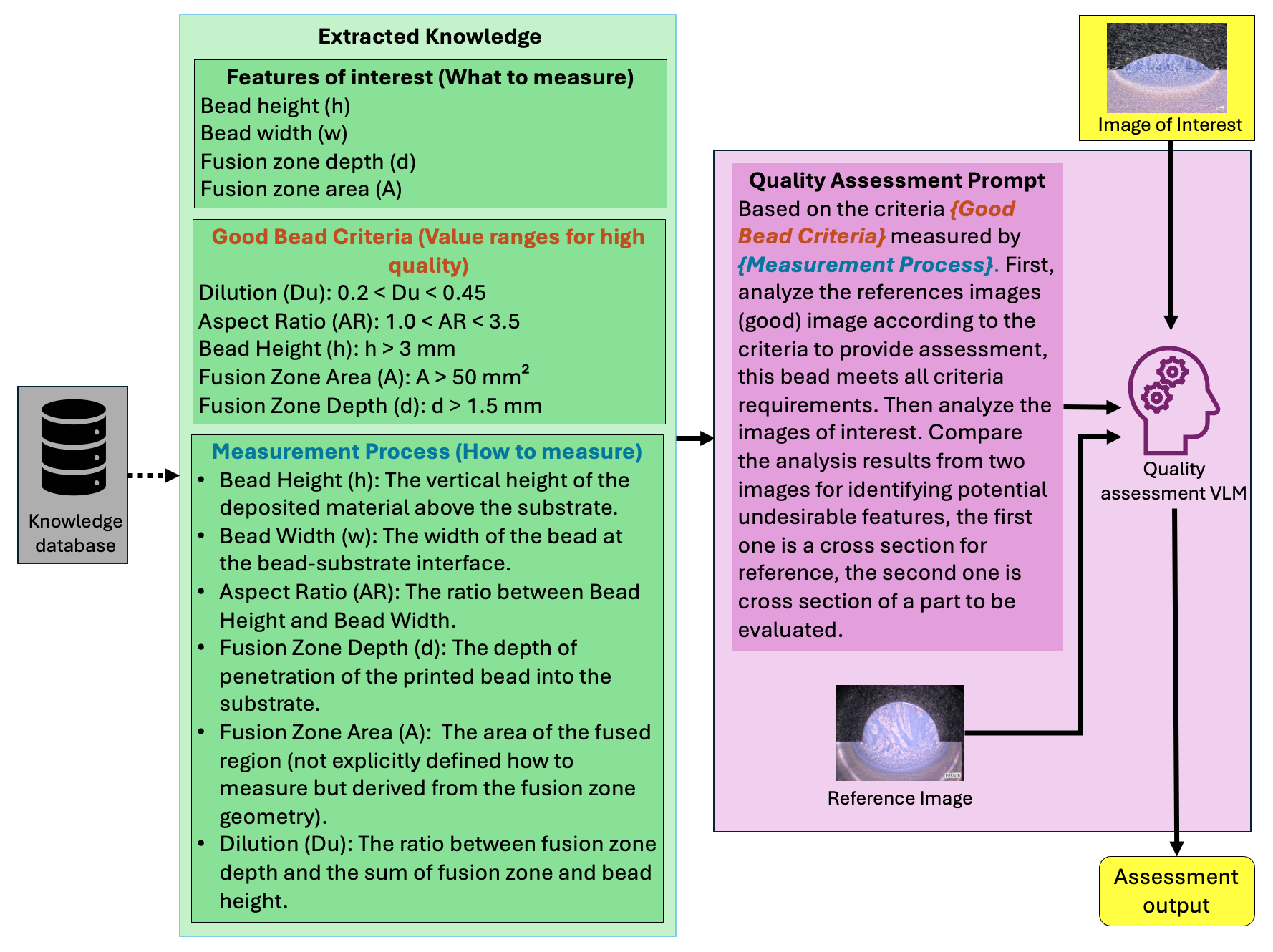}
    \caption{The knowledge-guided print quality assessment process with reference-guided knowledge grounding, the “Quality Assessment Prompt” and the reference images are the core components grounding process. This prompt first asks the print quality assessment VLM to get familiar with the knowledge by applying it to analyze a reference image capturing a print with the desired quality, then using its understanding to assess the actual image of interest.}
    \label{fig:Figure_5}
\end{figure}

To address this challenge, we introduce a reference-guided knowledge grounding approach. In this method, the VLM is guided through a step-by-step application of the evaluation criteria using an expert-selected, good-quality reference image. This process serves to align the VLM’s understanding of the quality assessment knowledge with visual context, enabling more accurate reasoning. Specifically, we decompose the expert reasoning process into a structured chain of prompts that reflect the decision-making steps a human expert would take when evaluating an image. The VLM is first asked to apply this reasoning chain to the reference image, which acts as a visual anchor for the learned criteria. One example prompt for the grounding is illustrated in the purple box in Figure~\ref{fig:Figure_5}.

Through this reference-guided knowledge grounding, VLMs can develop operational understanding of the measurement procedures and evaluation standards. Once this contextual alignment is achieved, the VLMs are then prompted to perform actual quality assessment tasks on target beads.

\section{Experiment Setup}
\subsection{Printing setup}
The quality assessment was conducted on 24 single-bead samples fabricated using Ti-6Al-4V wire feedstock. These samples were deliberately selected to provide a balanced representation of both high- and low-quality prints, enabling a fair evaluation of the model’s performance across diverse print conditions. While a larger dataset would offer a more comprehensive evaluation, the manual nature of the quality verification process, requiring expert review, posed a practical constraint on expanding the testing set. 

\begin{figure}
    \centering
    \includegraphics[width=1.0\linewidth]{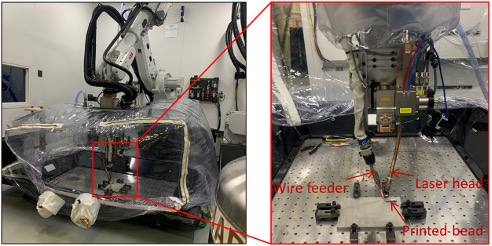}
    \caption{Wire-feed laser additive manufacturing setup, image from~\cite{Liu2022}. }
    \label{fig:Figure_6}
\end{figure}

\begin{figure}
    \centering
    \includegraphics[width=0.5\linewidth]{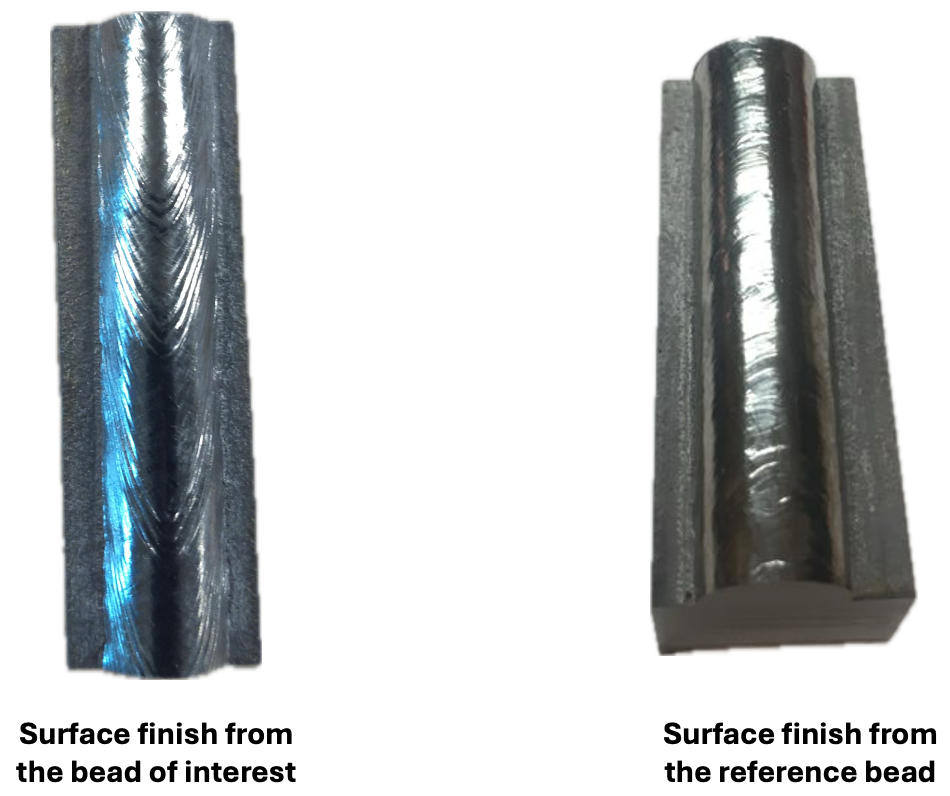}
    \caption{Print quality comparison between the bead of interest and the reference bead. The reference bead displays a smooth surface finish, while the bead of interest displays a rippled surface finish, indicating undesired print quality. Assessment models do not have access to the surface finish images.}
    \label{fig:Figure_7}
\end{figure}

The experiment setup is illustrated in Figure~\ref{fig:Figure_6}, the machine used is the same as the one in~\cite{Liu2022}, which is the one used for constructing the knowledge database. Using the same setup as in the knowledge database would reduce uncertainties in the manufacturing process, and help the downstream assessment focus only on the QA-VLM performance. A 6 kW laser head was employed to melt the wire, which was continuously supplied via a precision wire feeder. Both the laser head and the feeder were mounted on an industrial robotic arm to ensure precise control of the deposition path. The process was carried out within a sealed chamber filled with argon gas to maintain an inert atmosphere and prevent oxidation. Two examples of the produced beads are shown in Figure~\ref{fig:Figure_7}. The left bead shows some ripples on the surface and is considered a non-optimal or “bad” bead. The right bead has a continuous and smooth finish and is considered a “good” bead. The cross section of the smooth bead is used as the reference bead in the reference-guided knowledge grounding guidance process discussed in Section~\ref{knowledge grounding}.

\subsection{VLM assessment setup}

The journal article used for constructing the knowledge database is selected as~\cite{Liu2022}, which discussed correlations between the cross-sectional geometries to bead quality. It should be noted that although the QA-VLM structure can accept multiple journal articles to construct the knowledge database, we only used one for easy and clear downstream performance evaluations. 

The experiment involves two variations of the QA-VLM framework, each powered by a VLM with different sizes and reasoning capabilities. This design enables the evaluation of QA-VLM’s effectiveness across a range of model capacities. The selected VLMs are Gemini 2.0 Flash and Gemma-3 27B. Gemini, being the larger of the two, is expected to exhibit stronger visual understanding and reasoning abilities due to its greater scale. For text embeddings, we use the mxbai-embed-large model. Both the embedding model and Gemma-3 are hosted locally, while Gemini is accessed via Google’s infrastructure. To ensure deterministic and reproducible outputs, the temperature for all models is set to 0.

\subsection{Effectiveness evaluation with ablation study}
\begin{figure}
    \centering
    \includegraphics[width=1.0\linewidth]{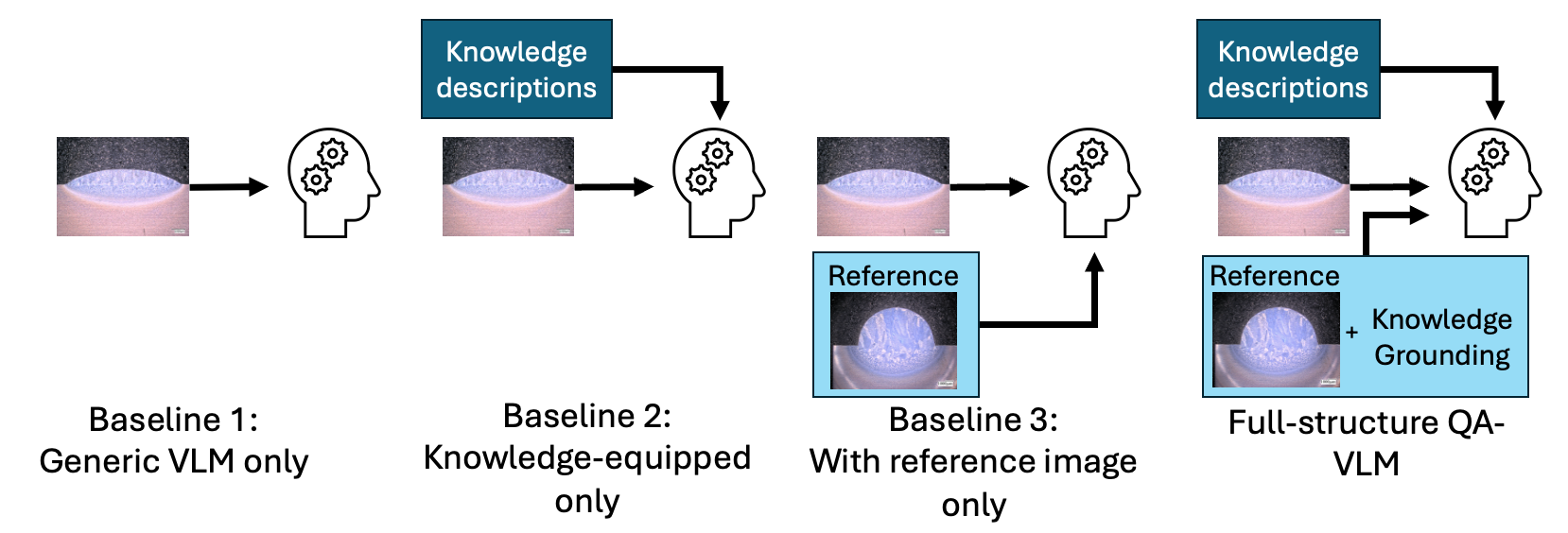}
    \caption{Three baseline models and the full-structure QA-VLM involved in the ablation study, it should be noted that adding knowledge grounding to baseline 3 configuration is not possible because the domain knowledge is not supplied.}
    \label{fig:Figure_8}
\end{figure}

To evaluate the contribution and effect of the individual and combined objectives on the quality assessment output, an ablation study is conducted by strategically leaving out certain objectives and performing the same quality assessment analysis. Differences in analysis indicate the influences of each objective on the evaluation output. The ablation study for is composed of 4 configurations, as shown in Figure~\ref{fig:Figure_8}: 1) generic, off-the-shelf VLM only, 2) knowledge-equipped VLM that only receives text descriptions of application-specific knowledge, 3)generic VLM that received the reference image in the reasoning chain process, and 4) the full-structure QA-VLM. It should be noted that although it would be ideal to keep the reasoning chain prompting for the baseline case 3 to study the influence of not providing the knowledge description, the absence of knowledge descriptions made the reasoning chain impossible, as the procedure that the reasoning chain was designed for depends on the knowledge.

Three key performance measurements: reasoning validity, knowledge relevance, and conclusion correctness, are used to evaluate each configuration. A well-performing model not only needs to provide correct conclusions backed by valid reasons, but it must also do so consistently with relevant knowledge. The definitions for reasoning validity, knowledge relevance, and conclusion correctness are provided below.

\begin{itemize}
    \item 	Reasoning validity measures the logical soundness of the assessments produced by the frameworks. A valid assessment should focus on relevant features and base its conclusions on accurate interpretations of those features. Any use of improper criteria or misinterpretation of key aspects, particularly when such errors lead to conclusions that contradict the actual condition of the same, makes an assessment output invalid. For example, when evaluating a bead with insufficient fusion zone depth, a valid model should correctly identify and assess the fusion zone rather than, say, mislabeling the substrate as part of it. Each response is reviewed by domain expert who determines whether the reasoning accurately reflects the physical condition of the printed part. The validity is a binary evaluation for each response, and the average score over the entire evaluation dataset is considered through Equation~\ref{eqn:eqn2}. In other words, any mistakes from the assessment model will make the response invalid. $N$ is the total number of samples in the testing set, $I$ is an indicator function, reflecting human expert judgment on the validity of the assessment output $y_i$

    \begin{equation} \label{eqn:eqn2}
    \text { Validity }=\frac{1}{N} \sum_{i=1}^N I\left(y_i\right)
    \end{equation} 

    \item Knowledge relevance measures both the appropriateness of the evaluation criteria and the consistency of the application. In this study, relevant knowledge centers around four core parameters: bead height, bead width, fusion zone depth, and fusion zone area, as outlined in~\cite{Li2022}. Derivative metrics such as aspect ratio and dilution are also considered acceptable since they are derived directly from the core parameters. To quantify this metric, we define a knowledge relevance score for each evaluation response as how many core parameters are discussed in the final response. Different from the validity measure, knowledge relevance will have integer scores between 0 and 4 for each assessment output rather than a binary measure. It should be noted that the assessment response can be valid while not achieving the full score for knowledge relevance in cases where the response only talks about the characteristics leading to low quality but misses the discussion on the remaining ones. The equation for knowledge relevance is illustrated in Equation~\ref{eqn:eqn3}

    \begin{equation} \label{eqn:eqn3}
        {\ R}_i=\max{\left(0,\ 4-E_i\right)}\ \     
    \end{equation}

    In Equation~\ref{eqn:eqn3}, $E_i$ is the total number of omitted relevant features and/or included irrelevant features in the evaluation sample $i$. The final score for a model is computed as the average over all test samples.

    \item Conclusion correctness refers to the accuracy of the final judgment rendered by the model. Regardless of how well a model reasons or applies knowledge. For instance, if a print bead meets all criteria and shows a smooth, well-fused profile, the model should correctly conclude that the print is of good quality. Any deviation from the correct outcome is recorded as an error in conclusion correctness. The conclusion correctness is also a binary measurement for each assessment and the average over all samples in the testing data represents model’s conclusion correctness performance. Since this metric does not depend on the preceding reasoning portion, it can be compared with other blackbox DL and ML methods. 

\end{itemize}

\section{Results and Discussion}
In this section, we first present comparisons of the models’ capabilities in providing valid and consistent answers, along with sample responses, and explore potential reasons for their behaviors. The models' overall performance across the total 24 test cases is then summarized in a table presented at the end of the section.

\subsection{Assessment validity}
\begin{figure}
    \centering
    \includegraphics[width=1.0\linewidth]{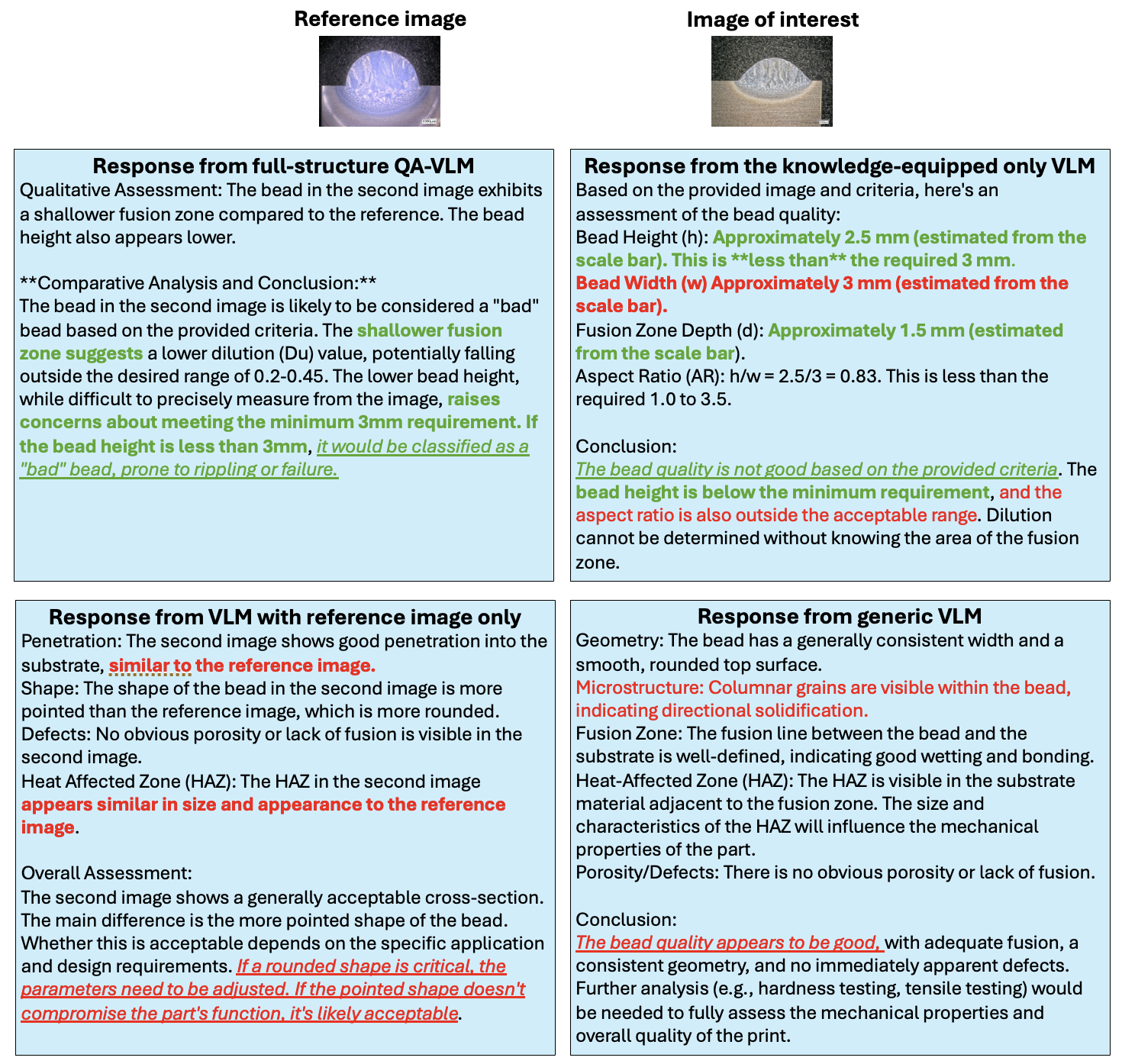}
    \caption{Validity comparison of the assessment comments made by QA-VLM and other baseline models. The response from the QA-VLM is shown in the top left corner. The three comparison baseline models involved are 1) the knowledge equipped only model that received text description of application-specific knowledge, 2) the VLM that only receives a good bead reference image, and 3) the generic VLM without any reference or knowledge, its responses are shown on the bottom left corner. The lack of relevant knowledge leads to invalid comments (bold red) and incorrect conclusions (red, underlined, italic) being drawn for the two setups without access to knowledge. Even with access to knowledge, VLM still cannot make fully correct responses without the help of knowledge grounding.}
    \label{fig:Figure_9}
\end{figure}

A sample output from all models involved in the ablation study is shown in Figure~\ref{fig:Figure_9}, the backbone VLM for these responses is Gemini. Among the four setups involved in the ablation study, the full-structure QA-VLM framework performed the best by stating factual observations about the bead dimension being too low (marked in gold green font), and drew the correct conclusion that the bead is not of good quality (marked in underlined green italic font). The surface finish of this bead is rippled due to low power and low wire feed rate, resulting in a relatively shallow fusion zone and low bead height. The full-structure QA-VLM successfully captures this characteristic and reflects such an observation in its quality assessment.  When only equipped with relevant knowledge, the Gemini VLM can correctly identify some feature measurements but makes errors on others: it misevaluated the bead width to be around 3 mm, while the actual width is measured to be around 5 mm. While the assessment conclusion for this bead is correct, stating that the bead is of low quality, one of the provided reasons is incorrect, which reduces the validity of this evaluation.

The two setups that do not have access to the relevant knowledge performed the worst. Their responses are shown in the bottom row. Because these setups lack knowledge of the proper procedures and metrics for quality assessment, much of their reasoning is flawed, resulting in incorrect conclusions. In cases where the model only has access to the reference image, the assessment VLM mistakenly assumes that the fusion zone of both beads is on the same level and concludes that the bead is of acceptable quality. In testing cases where neither a reference image nor knowledge is provided, the VLM relies purely on knowledge in its training data, which is mostly non-quantitative and human languages, causing its outputs to be mostly qualitative assessments without supportive analysis. Therefore, the produced assessment is inaccurate and ungrounded. Because neither of these models can provide correct conclusions supported by valid reasons, their reasoning validity is the lowest among the four testing cases.

\subsection{Assessment of knowledge relevance}
In addition to providing valid responses, a well-performing assessment model must consistently apply relevant domain knowledge in accordance with the assessment requirements. In the single-bead quality assessment application in this study, the relevant knowledge is about bead width, bead height, fusion zone depth, fusion zone area, and/or their derived features, according to the journal article that is used to construct the knowledge database. 

\begin{figure}
    \centering
    \includegraphics[width=1.0\linewidth]{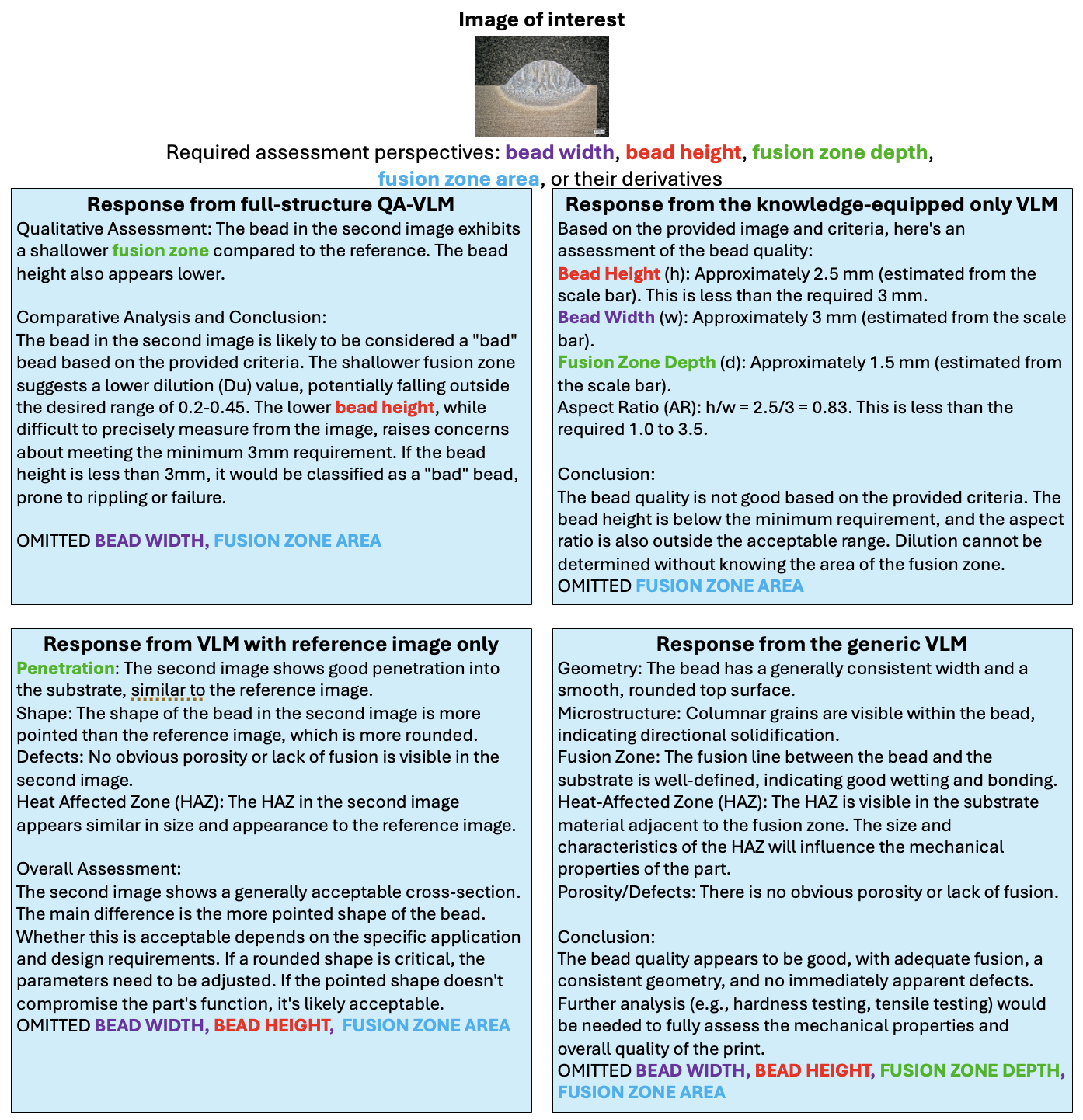}
    \caption{Knowledge relevance comparison of the assessment comments made by QA-VLM and other baseline models. Knowledge points are color-marked for knowledge relevance evaluation. The top two knowledge-equipped ones have the best knowledge relevance, while the bottom two that do not receive the knowledge have the lowest. The reference-guided knowledge grounding process caused drops in relevance; we think such guidance lets VLMs focus more on the areas where differences are the most significant and ignore the similar ones.}
    \label{fig:Figure_10}
\end{figure}

With the same response as used in Figure~\ref{fig:Figure_9}, we marked the key knowledge points used by each model in Figure~\ref{fig:Figure_10}. An ideal response would cover the evaluation from the perspectives of bead with, bead height, fusion zone depth, and fusion area. Two knowledge-equipped setups, the QA VLM and the knowledge-equipped only setup, demonstrated the highest levels of knowledge relevance by mentioning 2 and 3 features of interest (including derived features) without mentioning irrelevant features in their responses. Responses from these two models are shown in the top row in Figure~\ref{fig:Figure_10}.  The responses that mention key features are marked with non-black fonts. It is observed that, in general, the proposed framework shows a slight decrease in knowledge relevance due to some criteria omissions. We believe that this may be attributed to the knowledge grounding process with reference images that sometimes lead VLMs to put more attention on features that vary drastically and omit those that look similar. 

On the contrary, without guidance from application-specific knowledge, the knowledge relevance dropped drastically. These two non-knowledge-equipped models omitted most of the relevant knowledge and added some irrelevant or ambiguous criteria, such as shape, in their analysis response. Responses from these two models are shown in the bottom row in Figure~\ref{fig:Figure_10}. Such behavior is likely caused by their training dataset, where the analysis of similar images is performed from the perspectives discussed in the responses from these two models.  However, since the application and goal of the training material are different from the specific application analyzed in this paper, many of the features are irrelevant.  

\subsection{Overall evaluation}
In this section, we present the overall performance in validity and knowledge relevance with the summarized results collected from all 24 sample responses used in the ablation study. The results are summarized in Table~\ref{tab:Table_1}. A good assessment approach needs to have both high validity and knowledge relevance to present correct and human-interpretable answers. The introduced QA-VLM, powered by Gemini, achieved the highest overall performance. We attribute its success to both the access to related knowledge, the knowledge grounding, and the powerful image reasoning behaviors. The QA-VLM powered by Gemma also achieved significant improvements over its other setup variants, although the performance is not as good as that of the QA-VLM powered by Gemini. The slight performance drop could be attributed to the lower image reasoning capabilities in the Gemma model. 

\begin{landscape}
\begin{table}[htbp]
\centering
\footnotesize
\caption{Comparison of Angular Error, Average CPE for Evaluating Calibration}

\begin{tabular}{|c|c|c|c|c|c|c|c|c|c|}
\hline
\multirow{2}{*}{} & \multicolumn{4}{c|}{VLM: Gemma} & \multicolumn{4}{c|}{VLM: Gemini} & \multirow{2}{*}{{ML}} \\
\cline{2-9}
& QA-VLM & \makecell{Knowledge-\\equipped only} & \makecell{With\\reference only} & \makecell{Generic \\ VLM}
& QA-VLM & \makecell{Knowledge-\\equipped only} & \makecell{With\\reference only} & \makecell{Generic \\ VLM}& \\

\hline
Validity & 0.67 & 0.13 & 0 & 0.04 & \textbf{0.75} & 0.375 & 0.58 & 0.17 & NA \\

\hline
\makecell{Knowledge \\ relevance \\ (max 4)} & 3.54 & 3.63 & 0 & 0.04 & 3.63 & \textbf{3.96} & 0.88 & 0.08 & NA \\

\hline
\makecell{Conclusion\\correctness} & 0.67 & 0.58 & 0.54 & 0.5 & \textbf{0.95} & 0.58 & 0.83 & 0.54 & 0.74\\

\hline

\end{tabular}

\label{tab:Table_1}
\end{table}
\end{landscape}

When comparing validity, the smaller Gemma model benefits from the introduction of relative knowledge; compared with nearly all invalid answers produced by the two setups that do not have access to the knowledge, nearly 13\% of the knowledge-equipped Gemma VLM’s answers are valid responses. However, even with the validity increase, the validity score of 0.13 still shows that the knowledge-equipped Gemma cannot comprehend how to correctly apply the knowledge for quality assessment; knowledge application guidance via knowledge grounding is still required to significantly boost validity in the model’s response to around 0.67, as shown by the full-structure Gemma QA-VLM. Unlike the smaller Gemma model, Gemini can produce a fair number of valid responses without the help of relevant knowledge, as shown by the 0.17 score achieved by the generic Gemini model. This difference can be attributed to the rich pre-trained knowledge and strong reasoning capability in the off-the-shelf Gemini model. Interestingly, with the introduction of the application-specific knowledge, response validity from the Gemini model is lower than the configuration with reference image only, as demonstrated by the score of 0.375 and 0.58, respectively. We think this behavior may be attributed to the lack of understanding of how these text knowledge descriptions apply to evaluate quality in images. Once guided by the knowledge grounding process on how this knowledge applies, the Gemini model achieved the highest response validity of 0.75. 

When comparing knowledge relevance, similar behaviors can be observed across all four setups between the two VLMs; that is, the introduction of relevant knowledge significantly improves relevance. Responses from the two configurations without access to relevant knowledge contain nearly no evaluation from the perspective of key features of the bead, as shown by the low relevance score of 0.04 and 0.08 for Gemma and Gemini respectively. Their evaluation criteria seem to be consistently irrelevant and are potentially caused by the biases in their training data. Such behavior is especially obvious for the setups with reference image only powered by Gemma model, where all its responses are about porosity and lack of fusion, two behaviors not shown in the cross-section images. Among the two knowledge-equipped setups, the version without explicit knowledge grounding achieved the highest relevance scores, 3.63 for Gemma and 3.96 for Gemini, although the difference compared to the full-structure QA-VLM is small. We believe the decrease in knowledge relevance in the QA-VLM can be attributed to the model’s change in attention caused by the knowledge grounding. Prior to the guidance, VLMs may place equal attention on each feature and mention them equally in their responses. The guidance caused these models to focus more on the features that differ the most and less on those that look similar. 

The rend observed in conclusion correctness mirrors that of the validity comparison and can be attributed to the same underlying factors discussed in the validity section, that is, generic VLMs lack application-specific knowledge to draw correct conclusion, and the knowledge-equipping process effectively fills this gap. However, it is important to note that conclusion correctness focuses solely on whether the assessment of the print is accurate, regardless of the reasoning behind it. As a result, even responses deemed invalid can still yield correct conclusions. This explains why responses with near-zero validity can still achieve a conclusion correctness rate of around 50\%. Since ML models can only produce conclusions, they can only provide a performance comparison for conclusion correctness. 

\section{Conclusion}
In this work, we introduce a novel two-objective framework called QA-VLM that enables the production of human-interpretable quality assessment for DED-LW prints. The framework enables VLMs to make valid assessments by achieving two objectives: 1) equipping VLMs with application-specific knowledge, and 2) guiding the VLMs to understand how to apply the knowledge with the reference-guided knowledge grounding, with an image capturing high-quality print.  

Through an ablation study, we demonstrated the effects and benefits of achieving individual and combined objectives. In the application of assessing quality for the DED-LW bead, generic VLMs lack the application-specific knowledge and the know-how to correctly apply that knowledge. The realization of objective 1 allows VLMs to perform quality assessment from the perspective of application-specific knowledge, providing more relevant quality assessment outputs compared to generic, off-the-shelf VLMs. In the application to the quality assessment of single bead DED-LW, objective 1 enables VLMs to provide quantitative analysis on critical aspects such as bead height, bead width, fusion zone depth, and fusion zone area, which are indicative of bead quality, rather than irrelevant or ambiguous ones provided by generic VLMs. However, only achieving objective 1 does not guarantee the correct application of the knowledge. Therefore, objective 2 is designed to enable knowledge-equipped VLMs to learn how to apply with the guidance from exemplary images and knowledge grounding prompting. In the same quality assessment application, the VLM structure that achieves both objectives demonstrates higher assessment validity, as these models can apply the relevant knowledge correctly. Built on the application-specific and the know-how to apply it, the conclusion correctness of the quality assessment model is on par with existing machine learning models. Therefore, the QA-VLM framework can not only enable generic VLMs to achieve assessment correctness on par with existing ML models, but it can also provide a valid explanation to support the assessment, making the results more trustworthy.
% \end{linenumbers}


 \bibliographystyle{elsarticle-num} 
 \bibliography{QA-VLM_references}





% \end{thebibliography}
\end{document}